\documentclass[fleqn,10pt]{wlscirep}
\usepackage[utf8]{inputenc}
\usepackage[T1]{fontenc}
\usepackage{floatrow}
\usepackage{subfig}
\usepackage{algorithm}
\usepackage{algpseudocode}

\DeclareMathOperator*{\argmin}{arg\,min}
\DeclareMathOperator*{\minimize}{minimize}

\title{Forecasting high-dimensional dynamics exploiting suboptimal embeddings}

\author[1,2,*]{Shunya Okuno}
\author[1,3,+]{Kazuyuki Aihara}
\author[3,4,+]{Yoshito Hirata}
\affil[1]{Institute of Industrial Science, The University of Tokyo, 4-6-1 Komaba, Meguro-ku, Tokyo 153-8505, Japan}
\affil[2]{Disaster Reduction \& Environmental Engineering Department, Kozo Keikaku Engineering Inc., 4-5-3 Chuo, Nakanoku, Tokyo 164-0011, Japan}
\affil[3]{International Research Center for Neurointelligence (WPI-IRCN), The University of Tokyo, 7-3-1 Hongo, Bunkyo-ku, Tokyo 113-0033, Japan}
\affil[4]{Mathematics and Informatics Center, The University of Tokyo, 7-3-1 Hongo, Bunkyo-ku, Tokyo 113-0033, Japan}
\affil[*]{okuno@sat.t.u-tokyo.ac.jp}
\affil[+]{these authors contributed equally to this work}


\begin{abstract}
Delay embedding---a method for reconstructing dynamical systems by delay coordinates---is widely used to forecast nonlinear time series as a model-free approach.
When multivariate time series are observed, several existing frameworks can be applied to yield a single forecast combining multiple forecasts derived from various embeddings.
However, the performance of these frameworks is not always satisfactory because they randomly select embeddings or use brute force and do not consider the diversity of the embeddings to combine.
Herein, we develop a forecasting framework that overcomes these existing problems.
The framework exploits various ``suboptimal embeddings'' obtained by minimizing the in-sample error via combinatorial optimization.
The framework achieves the best results among existing frameworks for sample toy datasets and a real-world flood dataset.
We show that the framework is applicable to a wide range of data lengths and dimensions.
Therefore, the framework can be applied to various fields such as neuroscience, ecology, finance, fluid dynamics, weather, and disaster prevention.
\end{abstract}

\begin{document}
\flushbottom
\maketitle
\thispagestyle{empty}

\section*{Introduction}
Forecasting a future system state is an important task in various fields such as neuroscience, ecology, finance, fluid dynamics, weather, and disaster prevention. If accurate system equations are unknown but a time series is observed, we can use model-free forecasting approaches.
Although there are various model-free methods, if the time series is assumed to be nonlinear and deterministic, one of the most common approaches is delay embedding. According to embedding theorems\cite{Takens1981, Sauer1991a}, we can reconstruct the underlying dynamics by delay embedding (see Methods for details). These theorems ensure that the map from the original attractor to the reconstructed attractor has a one-to-one correspondence. These embedding theorems have also been extended to multivariate data and nonuniform embeddings\cite{Deyle2011}.

When multivariate time series are observed, various embeddings can be obtained on the basis of the extended embedding theorem\cite{Deyle2011}. Ye and Sugihara\cite{Ye2016} exploited this property; they developed multiview embedding (MVE), which combines multiple forecasts based on the top-performing multiple embeddings scored by the in-sample error. MVE can yield a more accurate forecast than the single best embedding, especially for short time series\cite{Ye2016}. Although MVE works well with low-dimensional data, it cannot be simply applied to high-dimensional data. This is because MVE requires forecasts derived from all possible embeddings, whose total number combinatorially increases with the number of variables.
Although we can approximate MVE by randomly chosen embeddings if the total number of possible embeddings is beyond the computable one\cite{Okuno2019}, it is difficult to find high-performance embeddings for high-dimensional data.
Recently, Ma et al.\cite{Ma2018a} presented an outstanding framework, randomly distributed embedding (RDE), to tackle these high-dimensional data. Their key idea is to combine forecasts yielded by randomly generated ``nondelay embeddings'' of target variables. These ``nondelay embeddings'' can also reconstruct the original state space for some cases and drastically reduce the possible number of embeddings. Ma et al.\cite{Ma2018a} also showed that small embedding dimensions worked fine, even for high-dimensional dynamics, and successfully forecasted short-term high-dimensional data using the RDE framework.
Although RDE showed outstanding results, there is room for improvement for some specific tasks. For example, ``nondelay embedding'' may overlook important pieces of information if delayed observations carry such pieces of information, e.g., the upstream river height for flood forecasting. A random distribution may also be a problem for the case where only partial variables are valid system variables; in this case, most of the random embeddings are invalid and not suitable to forecast. In addition, several important hyperparameters---the embedding dimension and the number of embeddings to combine---must be manually or empirically selected.
Although ensembles tend to yield better results with significant diversity among its members\cite{Sollich96, Kuncheva2003}, RDE only aggregates forecasts for a fixed embedding dimension with a fixed number of embeddings to combine, regardless of their forecast performance.

In this study, we propose another forecasting framework that overcomes the disadvantages of MVE and RDE. Our key idea is to prepare diverse ``suboptimal embeddings'' via combinatorial optimization.
We combine the optimal number of these embeddings to maximize the performance of the combined forecast.
In contrast to the existing frameworks, the suitable embedding dimensions and the number of embeddings to combine are automatically determined through this procedure.
Our proposed framework achieves better forecast performance than the existing frameworks for high-dimensional toy models and a real-world flood dataset.

\section*{Results}
\subsection*{Proposed forecasting framework}
\begin{figure}[ht]
    \centering
    \includegraphics[width=.8\linewidth]{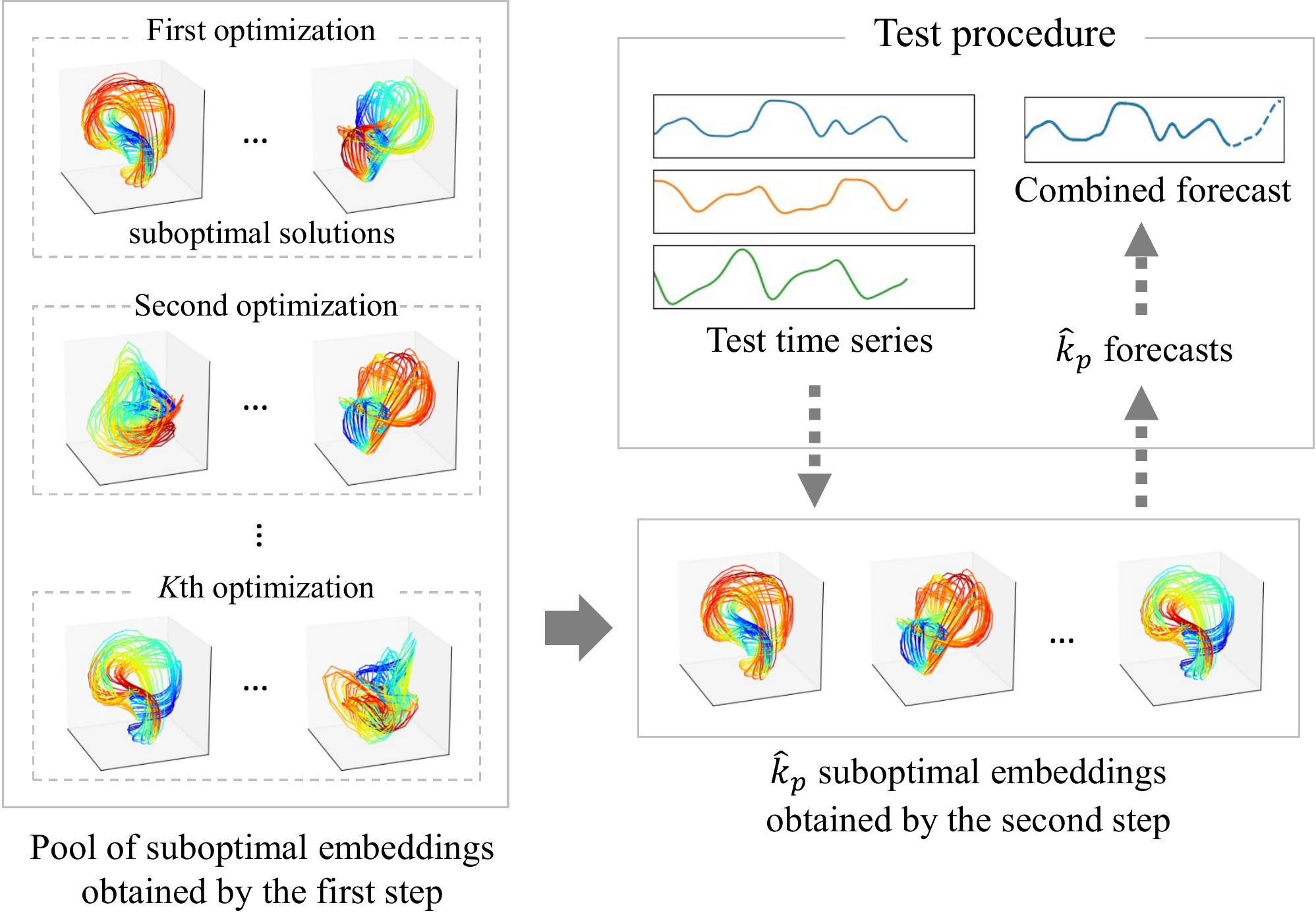}
    \caption{Schematic of the proposed forecasting procedure. We prepare a pool of suboptimal embeddings in the first step. We solve $K$ combinatorial optimization problems to obtain various embeddings in this step. Next, we pick $\hat{k}_p$ embeddings to minimize the error of the combined forecast in the second step. We combine the forecasts obtained by the $\hat{k}_p$ embeddings to test the time series.}
    \label{fig:procedure}
\end{figure}

Our proposed framework yields a single forecast to combine multiple ones, which are obtained by multiple embeddings. These embeddings are obtained through two steps.
First, we obtain suboptimal embeddings, which are suboptimal solutions to minimize the forecast error.
The aim of this procedure is to yield diverse embeddings to improve the performance of the combined forecast because such diversity is important for the performance of the ensemble\cite{Sollich96, Kuncheva2003}.
Second, we make the in-sample forecasts based on all suboptimal embeddings following MVE and RDE but pick the optimal number of embeddings to minimize the error of the combined forecasts. A schematic of the procedure is illustrated in Figure~\ref{fig:procedure}. We describe each step in detail in the following sub-subsections.

\subsubsection*{First step: Preparing suboptimal embeddings}
In our framework, we first obtain diverse suboptimal embeddings to minimize the in-sample forecast error. With the observed time series $y(t) \in \mathbb{R}^n$, a set of possible embeddings $\mathcal{E}$, and a corresponding delay vector $v_e (t) \in \mathbb{R}^E$ with $e \in \mathcal{E}$, we can obtain the $p$-steps-ahead forecast at time $t$ by a map $\psi_e$ as follows:
\begin{equation}
    \hat{y}_f^e (t+p|t) = \psi_e (v_e (t)).
\end{equation}
We can obtain such maps by suitable fitting algorithms, e.g., the conventional method of analogues\cite{Lorenz1969}, as in MVE, and Gaussian process regression\cite{Rasmussen2006}, as in RDE. Throughout this paper, we apply the same algorithm, a variation of the method of analogues (see Supplementary Information), to all forecasting frameworks including the proposed one. This is because we need to compare the performance of the frameworks on a fair basis that is not affected by the performance of each fitting algorithm.

To obtain high-performance embeddings in an efficient manner, we solve combinatorial optimization problems instead of random selection or using brute force.
We split the set of training time indices $\mathcal{T}_{train}=\{t \mid t < 0\}$ into $K$ datasets. Then, we solve the following optimization problem for each $K' \in \{1,2,...,K\}$ to minimize the error:
\begin{equation} \label{eq:fitness}
    \minimize_{e \in \mathcal{E}} \ \sum_{t \in \mathcal{T}_{train}^{K'}} \sum_p \|\hat{y}_f^e (t+p|t) - y_f (t+p)\|,
\end{equation}
where $\mathcal{T}_{train}^{K'}$ is the $K'$th set of training time indices and $\|\cdot\|$ is an appropriate norm. We use the $L^2$ norm throughout this paper. Note that $\hat{y}_f^e$ is computed in an in-sample manner; we use all of $\mathcal{T}_{train}$, except for the current time, as its library.
When we consider embeddings up to $l$ lags to embed the latest observation of the target variable, the number of possible embeddings is $2^{nl-1}$. If $nl$ is sufficiently large, it is almost impossible to compute all possible embeddings because of combinatorial explosion.
Here, we take a straightforward approach to minimize the forecast error: an evolution strategy that interprets an embedding as a one-dimensional binary series\cite{Vitrano2001, Small2003}. See Methods for details. Throughout this paper, we applied the $(\mu+\lambda)$ evolution strategy\cite{Schwefel1981a}.

In the process of each optimization, we store not only the best embedding but also the ``Hall of Fame,'' which preserves the best individuals in each generation. From the Hall of Fame, we select the best $M$ solutions that satisfy the condition that the Hamming distance $d(\cdot,\cdot)$ is larger than or equal to a certain threshold $\theta$ as follows:
\begin{equation} \label{eq:embedding_condition}
    d(e_i, e_j) \geq \theta, \ \forall i,j; i \neq j,
\end{equation}
where $e_i$ is the $i$th embedding of the sorted Hall of Fame.
Namely, we sort the Hall of Fame by the fitness (the objective function of equation \ref{eq:fitness}) and pick the best embeddings in order from the sorted Hall of Fame to satisfy equation~(\ref{eq:embedding_condition}).
This condition prunes similar embeddings and keeps diverse embeddings for the next step.
To apply this procedure, we obtain $K \cdot M$ suboptimal embeddings in total.
See also the first part of Algorithm~\ref{alg:framework} in Methods.

The procedure of the $K$ times optimizations yields diverse but high-performance embeddings.
If a dataset contains a large number of variables, it is extremely difficult to obtain the exact solution by evolution strategies or any other metaheuristic algorithms.
These algorithms find suboptimal solutions for most cases, and these solutions can be easily changed by slight changes in the optimization conditions.
The procedure rather exploits this property; the $K$ times optimization can find $K$ sets of diverse suboptimal solutions.
Instead of minimizing the whole in-sample error with $K$ different conditions, we minimize each $K$ split dataset and utilize the history of solutions to significantly reduce the computational time.
We numerically show the effect of $K$ times optimization in Supplementary Information.

To create diverse embeddings and improve the performance of the combined forecast, we can optionally consider $L$ finite impulse response (FIR) filters for each suboptimal embedding; namely, $K \cdot L \cdot M$ embeddings are obtained in total. These filtered embeddings are justified by the Filtered Delay Embedding Prevalence Theorem\cite{Sauer1991a}, and a study has shown that the combination of various forecasts via linear filters improves the forecast result in some cases\cite{Okuno2017a}. Although we can compute suboptimal embeddings for each filter, we simply consider $L$ filters based on the obtained $K \cdot M$ embeddings. Empirically, the optimization of the embeddings for each filter does not realize a significant improvement in performance to justify the additional computational cost.
See Methods for the details of forecasting filtered time series.
In this paper, we set the filter coefficients as $h_i(0)=1, h_i(1)=\rho \ \forall i$ for all numerical examples.


\subsubsection*{Second step: Combining forecasts to minimize the in-sample error}
Next, we pick the optimal number of embeddings from the $K \cdot L \cdot M$ embeddings to minimize the error of the combined forecast.
We compute in-sample forecasts by the embeddings for all of $\mathcal{T}_{train}$, as done in the MVE or RDE aggregation schemes.
We determine the optimal number of embeddings to integrate forecasts, which is neither the square root of the possible number of embeddings (as MVE does) nor the embedding dimension (as the RDE aggregation scheme does).
Here, we define a tuple of the forecast indices sorted by the in-sample error for step $p$ as $\mathcal{I}_p$, and the $i$th best embedding is written as $\mathcal{I}_p(i)$. Note that $\mathcal{I}_p$ is separately computed for each forecast horizon because accurate embeddings may differ according to the forecast horizons, as discussed later. We obtain $\hat{k}_p$ to minimize the combined forecast as follows:
\begin{align}
    Y_k(t+p|t) := 1/k \sum_{i=1}^k \hat{y}_f^{\mathcal{I}_p(i)} (t+p|t), \\
    \hat{k}_p = \argmin_{k=1,2,...,P} \sum_{t \in \mathcal{T}_{train}} [Y_k(t+p|t) - y_f(t)]^2, \\
    \hat{y}_f(t+p|t) = Y_{\hat{k}_p}(t+p|t).
\end{align}
We can obtain $\hat{k}_p$ in a brute-force manner with a very small computational cost because the $k$th combined forecast $Y_k(t+p|t)$ can be recursively computed as follows:
\begin{equation}
    Y_{k+1}(t+p|t) = \left[k Y_k(t+p|t) + \hat{y}_f^{\mathcal{I}_p(k+1)}(t+p|t) \right] / (k+1),
\end{equation}
where $k=1,2,...,P-1,\  Y_1(t+p|t) = \hat{y}_f^{\mathcal{I}_p(1)}(t+p|t)$. We compute $Y_k$ and the corresponding error for all $k$ and pick the optimal value to minimize the squared error. Note that $\hat{k}_p$ can be different for each $p$; empirically, $\hat{k}_p$ tends to be larger for a longer step.

For unobserved samples $t \geq 0$, the forecast $\hat{y}_f (t+p|t)$ is computed as follows:
\begin{equation}
    \hat{y}_f (t+p|t) =  \sum_{k=1}^{\hat{k}_p} \hat{y}_f^{\mathcal{I}_p(k)}(t+p|t) / \hat{k}_p.
\end{equation}
Note that only $\hat{k}_p$ forecasts need to be computed rather than $K \cdot L \cdot M$ forecasts.
We summarize the whole algorithm in Algorithm~\ref{alg:framework} in Methods.

\subsection*{Numerical experiments}
\subsubsection*{Toy models}
We demonstrate our framework with toy models.
We first tested with the 10-dimensional Lorenz'96I model \cite{Lorenz1995a}. The model is described as 10-dimensional differential equations, and we used five of their system variables and five random walks as a dataset---namely, Lorenz'96I's $x_0, x_1, x_2, x_3, x_4$ and random walks $x_5, x_6, x_7, x_8, x_9$. We forecasted $x_0$ up to 10 steps.
We compared the proposed framework with existing frameworks---namely, randomly distributed embedding with an aggregation scheme (RDE), multiview embedding (MVE), a variation of MVE (state-dependent weighting\cite{Okuno2019} (SDW)), and the single best embedding via the $(\mu+\lambda)$-ES algorithm. See Supplementary Information for the detailed conditions.

\begin{figure}[]
    \centering
    \includegraphics[width=1.\linewidth]{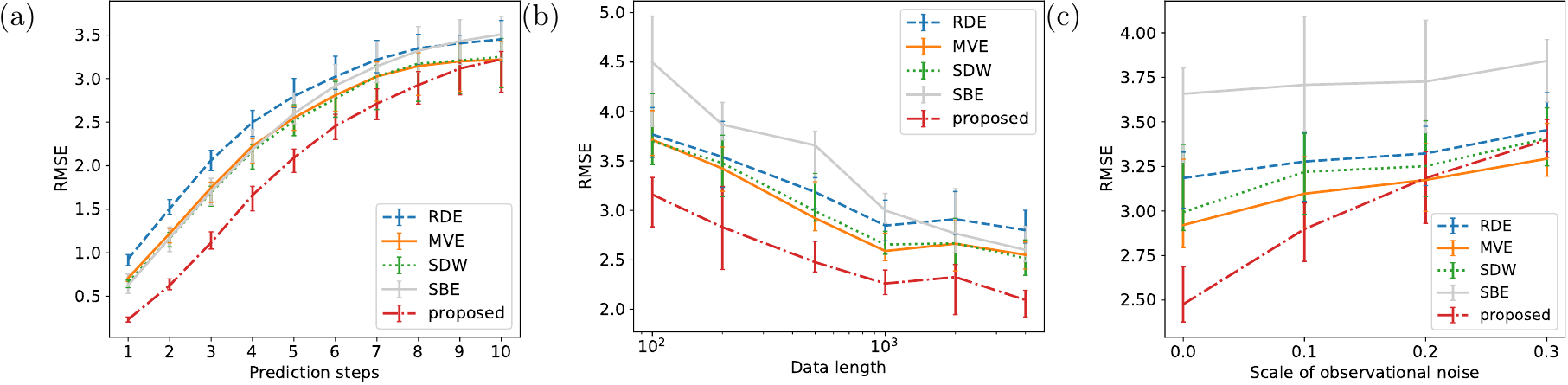}
    \caption{Forecast performance with the Lorenz'96I model: comparisons of the performance (a) up to 10 steps ahead with the fixed data length (4000) without noise, (b) with different values of the data length, and (c) with different scales of observational noise. We computed the RMSE of the five-steps-ahead forecasts with randomly distributed embedding (RDE), multiview embedding (MVE), state-dependent weighting (SDW), single-best embedding based on the $(\mu+\lambda)$-ES algorithm (SBE), and the proposed framework. These tests were carried out with 20 datasets generated with different random initial conditions and noise. The median, upper quartile, and lower quartile are shown.}
    \label{fig:lorenz96_rmse}
\end{figure}

We first tested the performance with the data length of 4000 for the database and 500 for the evaluation. Figure~\ref{fig:lorenz96_rmse}(a) shows that the proposed forecast achieved the best performance among the existing frameworks up to 10 steps ahead with the length of 4000.

We conducted a sensitivity analysis with the data length. We set the data length, i.e., the size of the training dataset, to \{100, 200, 500, 1000, 2000, 4000\} and tested the performance with 500 samples for all cases. Figure~\ref{fig:lorenz96_rmse}(b) shows that the proposed five-steps-ahead forecast yielded the minimum error among the existing frameworks for all the data lengths.

We also checked the robustness against observational noise. We added Gaussian observational noise scaled by the standard deviation of the target variable and tested the performance with a scale range of \{0.0, 0.1, 0.2, 0.3\}. We set the training data length as 500 and evaluated with 500 samples.
Figure~\ref{fig:lorenz96_rmse}(c) shows that a lower noise level results in better performance for the proposed framework compared with the other frameworks. Although the proposed framework achieved the best performance up to 10\% noise, the performance was degraded by high-level noise and is the second best for 30\% noise.

\begin{figure}[h]
    \centering
    \includegraphics[width=1.\linewidth]{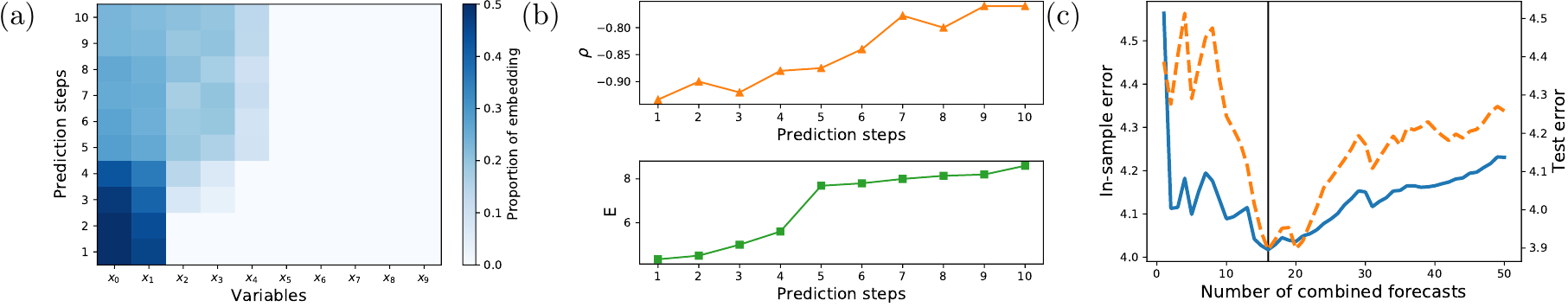}
    \caption{Forecast profile of the Lorenz'96I model with the data length of 4000. Panel (a) shows the proportion of embedding of the proposed forecasts. The color indicates the proportion of embedding for each variable averaged over the number of combined forecasts for each step. Note that for each prediction step, the sum of the proportions of all variables is one. Variables $x_0,...,x_4$ are the variables of the 10-dimensional Lorenz'96I equations, and the other variables are random walks. Panel (b) shows the averaged filter coefficient $\rho$ and embedding dimension $E$ over the combined forecasts for each step. Note that the averaged dimensions can be a decimal fraction since they are averaged over the number of combined forecasts for each step. Panel (c) shows the relation between the number of combined forecasts and the errors. The solid line shows the in-sample error, and the dashed line shows the test error. The vertical solid line shows the selected number of combined forecasts $k_p$ to minimize the in-sample error.}
    \label{fig:lorenz96_profile}
\end{figure}

We profiled which variables were embedded for the combined forecasts for the case where the data length is 4000 without noise. We summed $\hat{k}_p$ embeddings for each variable, where 1 means embedded and 0 means not embedded. Then, we computed the proportion of embedding of each variable, normalizing the sum to one for each $p$. Figure~\ref{fig:lorenz96_profile}(a) shows that random walks $x_5, x_6, x_7, x_8$, and $x_9$ were seldom embedded with the proposed framework. This result suggests that our proposed framework can select the appropriate variables. The results also show that the proportion of embedding of the target variable $x_0$ is large for short-term prediction and small for longer-term prediction. This result is consistent with that of an existing study\cite{Chayama2016a}.

We also profiled the filter coefficient $\rho$ and the embedding dimension $E$ of the proposed combined forecasts.
We averaged $\rho$ and $E$ over the $\hat{k}_p$ combined members for each step. Figure~\ref{fig:lorenz96_profile}(b) shows that smaller dimensions are selected for shorter steps, and larger dimensions are selected for longer predictions. The figure also shows that smaller and larger values of $\rho$ are selected for shorter and longer steps, respectively.
Because we employ high-pass filtering $(\rho\leq0)$ with this model, a smaller $\rho$ focuses on a smaller time period and vice versa.

We investigated the number of combined forecasts and the forecast errors to check whether the proposed integration scheme works properly. Although the proposed framework determines $\hat{k}_p$ on the basis of the in-sample error, $\hat{k}_p$ also minimizes the test error in this example (see Figure~\ref{fig:lorenz96_profile}(c)).

We also demonstrate our framework with the Kuramoto--Sivashinsky equations\cite{Kuramoto1976, Sivashinsky1977}. We generated 20-dimensional time series: the values of the first 10 grids of the Kuramoto--Sivashinsky equations $x_0, x_1, ..., x_9$ and 10 random walks $x_{10}, x_{11}, ..., x_{19}$. We forecasted $x_0$ up to 10 steps using the same parameter values, data length, and noise scales as used in the Lorenz'96I example. See Supplementary Information for the detailed conditions.

\begin{figure}[]
    \centering
    \includegraphics[width=1.\linewidth]{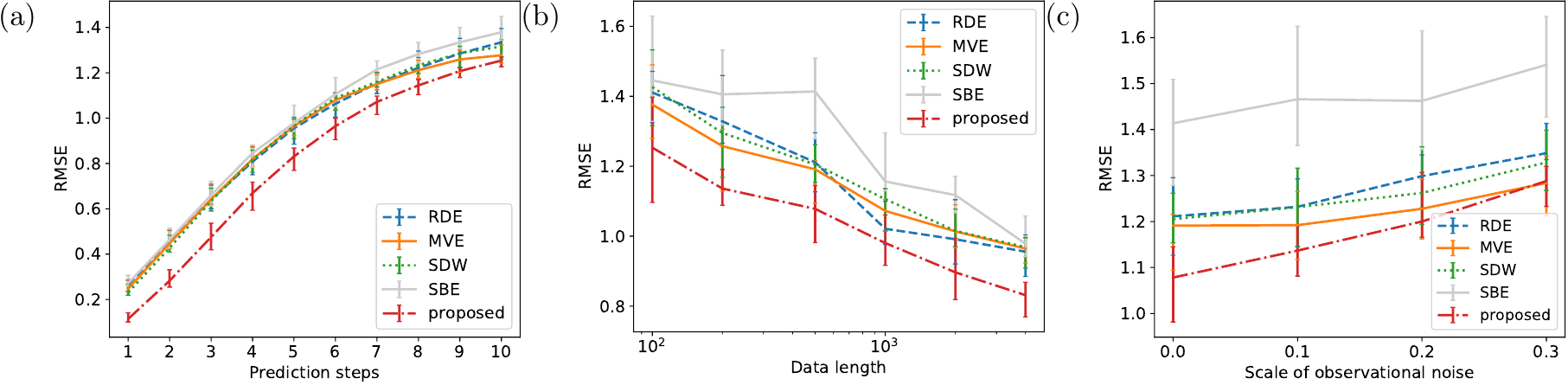}
    \caption{Forecast performance with the Kuramoto--Sivashinsky equations: comparisons of the performance (a) up to 10 steps ahead with the fixed data length (4000) without noise, (b) with different values of the data length, and (c) with different scales of observational noise. We computed the RMSE of the five-steps-ahead forecasts with randomly distributed embedding (RDE), multiview embedding (MVE), state-dependent weighting (SDW), single-best embedding based on the $(\mu+\lambda)$-ES algorithm (SBE), and the proposed framework. These tests were carried out with 20 datasets generated with different random initial conditions and noise. The median, upper quartile, and lower quartile are shown.}
    \label{fig:kuramoto_rmse}
\end{figure}

Despite the increases in the variables, Figures~\ref{fig:kuramoto_rmse}(a) and (b) show that the proposed forecast yielded the minimum error among the existing frameworks, as in the Lorenz'96I example. Regarding the noise sensitivity, a lower noise level results in better performance for the proposed framework compared to the other frameworks. The proposed framework achieved the best performance for a wide data range and relatively small observational noise.

\begin{figure}[]
    \centering
    \includegraphics[width=1.\linewidth]{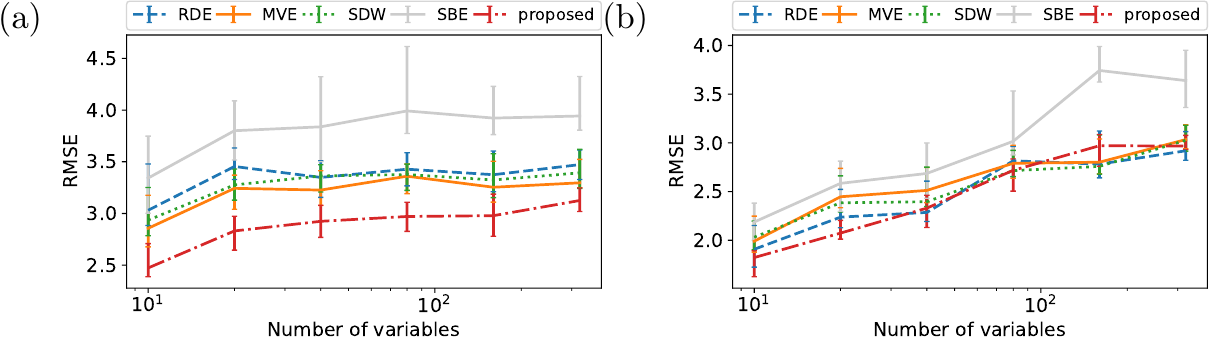}
    \caption{Forecast performance with the Lorenz'96I model for various numbers of variables: cases where (a) a half of the variables are substituted with random walks and (b) all variables are available. We computed the RMSE of the five-steps-ahead forecasts with randomly distributed embedding (RDE), multiview embedding (MVE), state-dependent weighting (SDW), single-best embedding based on the $(\mu+\lambda)$-ES algorithm (SBE), and the proposed framework. These tests were carried out with 20 datasets generated with different random initial conditions and noise. The median, upper quartile, and lower quartile are shown.}
    \label{fig:lorenz96_variables}
\end{figure}

Finally, we tested the sensitivity to the number of variables. We changed the dimensions of Lorenz'96I to \{10, 20, 40, 80, 160, 320\}, and we substituted a half of each dataset with random walks. We set the training data length to 500 and evaluated with 500 samples.
For all ranges of variables, the proposed framework achieved the best performance (Figure~\ref{fig:lorenz96_variables}(a)).
We also tested the case where all variables are available and the datasets do not contain random walks; this is the case in which delay embedding is not needed because all system variables are observed.
In this case, the performance difference is negligible for all frameworks (except the single best embedding), and RDE achieved the best score for the case where the number of variables is 320.
Although this condition is too ideal because the observation function is an identity map, we can expect RDE to perform well for such cases that include many informative variables.

\subsubsection*{Flood dataset}
We tested the proposed framework with real-world data, namely the flood forecasting competition dataset at ``Artificial Neural Network Experiment (ANNEX 2005/2006)''\cite{Dawson2005a}. This dataset contains nine variables: the river stages of the target site and three upstream sites and five rain gauges for three periods. We forecasted 6, 12, 18, and 24 h ahead of the target river stage using all nine variables. See Supplementary Information for the detailed conditions.
We also compared our framework with four other conventional machine-learning methods used in an existing study\cite{Okuno2019}: a recurrent neural network with long short-term memory (LSTM)\cite{Hochreiter:1997:LSM:1246443.1246450}, support vector regression (SVR)\cite{Boser:1992:TAO:130385.130401}, and random forest regression\cite{Breiman2001}.

\begin{figure}[htb]
    \begin{center}
        \includegraphics[width=.9\linewidth]{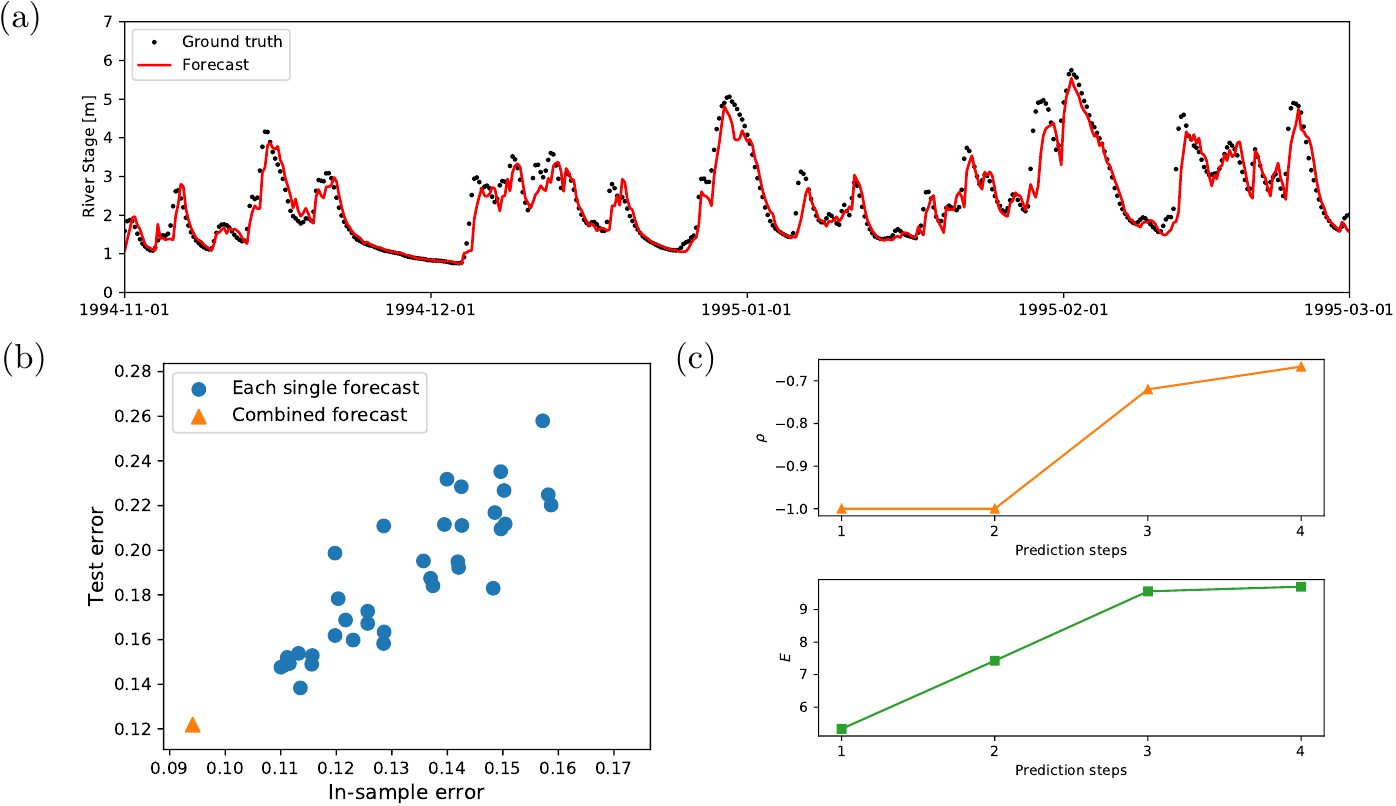}
        \caption{Forecast results for the flood dataset. Panel (a) shows a comparison of the ground truth and the proposed 24-h-ahead forecast. The proposed forecast did not underestimate the maximum river stage, which is the maximum value of the whole dataset. Panel (b) shows a comparison of the in-sample and test errors of the ensemble members. The results demonstrate the difficulty of selecting the best forecast because the best in-sample forecast does not always perform the best. In contrast, the combined forecast performed the best by far for both training and test data. Panel (c) shows the filter coefficient $\rho$ and embedding dimension $E$ averaged over the combined forecasts for each step.}
        \label{fig:flood}
    \end{center}
\end{figure}

The results in Table~\ref{tab:flood} show that the proposed framework yielded the best root-mean-square error (RMSE) through 12--24 h ahead, and the differences compared to the other methods were expanded over the forecast horizon.
Although the 6 h (one step)-ahead forecast was not the best, the difference compared to the best score was negligible. The dataset contains only 1460 points, and long-term memory is not required for this task; hence, LSTM did not perform well.
Although the test data include the largest river stage on 1995-02-01, Figure~\ref{fig:flood}(a) shows that the proposed framework properly forecasted the inexperienced river stage.

We compared the combined forecast with its individual members obtained with the $(\mu+\lambda)$-ES algorithm. Regarding the individual members, the order of the in-sample score was different from the order of the test (see Figure~\ref{fig:flood}(b)). In contrast, the combined forecast achieved the minimum error by far for the both the in-sample and test errors. This forecast stability is one of the largest advantages of combining forecasts.

We also profiled $\rho$ and $E$ averaged over the $\hat{k}_p$ combined members for each step. The results in Figure~\ref{fig:flood}(c) suggest the same trends as the toy models, even for the real-world data. Specifically, smaller dimensions and smaller values of $\rho$ were selected for shorter steps.

\section*{Discussion}
Our numerical experiments suggest that our proposed framework works well for a wide range of the data length, including short time series such as 100. In contrast, the framework did not appear to work fine with very noisy data, and MVE achieved the best performance, as shown in the noise sensitivity analyses (Figures~\ref{fig:lorenz96_rmse}(c) and ~\ref{fig:kuramoto_rmse}(c)). This suggests that noise prevents the framework from selecting optimal embeddings on the basis of the squared errors, and it results in obtaining invalid embeddings. MVE can still be a good option for very noisy data.
Note that the performance of RDE was not satisfactory in some of our numerical experiments, but this is because the RDE framework was originally proposed for extremely high-dimensional data, e.g., hundreds of variables. As shown in Figure~\ref{fig:lorenz96_variables}(b), we can expect RDE to perform well for cases that include many informative variables, and delay embedding is not needed.

As shown in the numerical experiments, the proposed framework worked well for high-dimensional dynamics and datasets containing high-dimensional variables. On the other hand, it is not suitable to apply it to relatively low-dimensional data. We tested the framework with low-dimensional datasets, and the results suggest that the proposed framework does not always yield the best performance, especially for low-dimensional datasets such as the R\"ossler equations\cite{Rossler1976} (see Supplementary Information).
This is because it is easy to find suitable embeddings to forecast in a brute-force manner, especially for low-dimensional datasets, while it becomes combinatorially difficult as the number of variables increases.
In addition, the aim of $K$ times optimization is to yield diverse embeddings (see the effect of $K$ times optimization in Supplementary Information).
However, if a dataset contains a small number of variables, almost the exact solutions are obtained for every optimization.

The proposed method combines forecasts via the simple average, optimizing the number of forecasts. The simple average works fine for most cases because suboptimal embeddings can equally forecast well thanks to solving the combinatorial optimization problem, and the test performance order is not always the same as the corresponding training data, especially for real-world data (see Figure~\ref{fig:flood}(b)).
If suboptimal embeddings show quite different forecast performances, the weighted average based on the error distribution\cite{Ma2018a} may work better. We can also apply the expert advice framework\cite{Cesa-Bianchi2006a} if better forecasts are assumed to be changed over time.

We need to select the values of several parameters to apply the proposed framework, i.e., the number of split datasets $K$, the number of the suboptimal embeddings for each batch $M$, the criteria of the Hamming distance $\theta$, the number of filters $L$, and their components.
We can control the variations in forecasts with these parameters. In order to set $K$, we consider the number of samples in each divided dataset; empirically, the framework works well with more than 50 samples for each divided dataset. Samples that are too small prevent the framework from finding good embeddings, but samples that are too large reduce the diversity of the embeddings.
Although $M$ can be any large integer (up to the total number of individuals of the evolution strategy), $M\leq5$ is sufficient for most cases. Note that the value of $M$ that is too large is overkill and unnecessarily increases the computational cost.
The criteria for the Hamming distance $\theta$ directly controls the similarity of suboptimal embeddings. We set $\theta=3$ in this study, and it worked fine for all cases. Note that conventional MVE and RDE guarantee that the Hamming distance among all embeddings is larger than two (i.e., $\theta=2$) for a fixed embedding dimension.
In practice, appropriate filters depend on the prediction steps. If we predict shorter time periods, high-pass filters will work fine; if we predict longer time periods, low frequencies are needed to obtain fine forecasts, as the numerical experiments suggest. Although we treat high-pass filters in this study, low-pass filters may work for extremely noisy data.
We need to understand the effects of these parameters and set the appropriate values within reason, $P=KLM$, which is usually much smaller than the total number of embeddings evaluated in the first step.

Our numerical experiments suggest that the appropriate embeddings vary with the prediction steps. The results show that a smaller $E$ with a smaller $\rho$ is preferable for a short-term forecast, and a larger $E$ with a larger $\rho$ is preferable for a longer-term forecast. These results are consistent with those of an existing study\cite{Chayama2016a} and with the fact that the longer-step forecasts need more complex map from the current inputs to the forecasts. Therefore, more information is needed to express the complex map.

Although we applied the $(\mu+\lambda)$-ES algorithm to obtain the suboptimal embeddings, it is worth considering other methods to solve the combinatorial optimization problem. We can apply other evolution algorithms, e.g., simulated annealing\cite{Kirkpatrick83optimizationby} and ant colony optimization\cite{Dorigo:2004:ACO:975277}. Any algorithm may work as long as it can store the suboptimal embeddings through its optimization procedure. Testing and evaluating other algorithms are open problems.

Throughout this study, we used a variation of a local nonlinear prediction method to forecast the target variable. One of the most significant advantages of these local prediction methods is that we can compute in-sample forecasts very easily with a small computational cost; all we have to do is to exclude the current query from the nearest neighbors.
In addition, owing to recent progress in approximate neighboring search algorithms (e.g., Refs.~\citenum{Muja2014a, Fu2016a}), these algorithms are fast enough to apply the evolution algorithms. Moreover, it is also suitable to obtain suboptimal embeddings through the forecast errors because local nonlinear prediction methods are sensitive to the false nearest neighbors derived from invalid features.
However, we can apply other regression methods, e.g., Gaussian process regression\cite{Rasmussen2006}, as done in RDE\cite{Ma2018a}, for further improvements. Note that these methods incur a very large computational cost to yield in-sample forecasts, as mentioned in Ref.~\citenum{Ma2018a}. To reduce the computational cost, we can apply these methods to only the process in the second step and can approximate the in-sample error by the k-fold cross-validation error.

\section*{Methods}
\subsection*{Delay embedding}
Delay embedding is a method of reconstructing the original state space using the delay coordinates $v(t)$. Here, with the $n$-dimensional observed time series $y(t)=[y_1(t),y_2(t),...,y_n(t)]$, variable $\sigma_i \in \{1,2,...,n\}$, and time lag $\tau_i \in \{0,1,...,l-1\}$, $v(t) \in \mathbb{R}^E$ is defined as follows:
\begin{equation}
    v(t)=\left[ y_{\sigma_1}(t-\tau_1), y_{\sigma_2}(t-\tau_2),...,y_{\sigma_E}(t-\tau_E) \right],
\end{equation}
where $E$ is the embedding dimension, $\tau_j=0 \ \exists j \in \{1,2,...,n\}$, and no duplication is allowed for any element. According to embedding theorems\cite{Takens1981,Deyle2011}, an attractor reconstructed by $v(t)$ is an embedding with the appropriate $E$. Note that the number of possible reconstructions increase combinatorially.

\subsection*{Optimizing embedding by an evolution strategy}
One can optimize embedding to solve combinatorial optimization problems to treat embeddings as binary series\cite{Vitrano2001, Small2003}.
Here, we solve equation~(\ref{eq:fitness}) by a combinatorial optimization problem.
We treat embeddings as binary series, i.e., 1 meaning embedded and 0 meaning not embedded.
For example, with $n=3$ and $l=2$, the embedding $[y_1 (t),y_2 (t),y_3 (t-1)]$ can be expressed by the $nl=6$ binary code $e=[1,0,1,0,0,1]$: the first element 1 corresponds to $y_1 (t)$ to embed, the second element 0 corresponds to $y_1 (t-1)$ to embed, the third element 1 corresponds to $y_2 (t)$ to embed, and so on.
The optimization of a binary series is a typical combinatorial optimization problem.
In this paper, we apply the $(\mu+\lambda)$ evolution strategy\cite{Schwefel1981a}.
The parameter $\mu$ denotes the number of parents, $\lambda$ denotes the number of offspring, and $+$ denotes {\it plus selection}, which means that the next parents are selected from both the previous parents and offspring.
Precisely, the algorithm yields $\lambda$ new offspring from the top-performing $\mu$ individuals selected from the previous parents and offspring.
The algorithm is an elitist method since the algorithm retains the best individuals unless they are replaced by superior individuals. The algorithm tends to converge fast, but it is likely to become trapped at local optima. Therefore, the algorithm is suitable for our proposed framework.
However, it is worth considering the application of other existing methods\cite{Runge2015, Vlachos2009, Chen2010} to obtain suboptimal embeddings.

\subsection*{Forecasting through FIR filters}
We apply FIR filters to the original time series to obtain various embeddings and forecasts. We apply the following linear transformation for each variable:
\begin{equation}
    z_i(t) = \left[\sum_{k=0}^{N-1} h_i(k) y_i(t-k) - \mu_i \right] / \sigma_i,
\end{equation}
where $h_i$'s are filter coefficients with $N$ elements, $h_i(0) = 1 \ \forall i$, and $\mu_i$ and $\sigma_i$ are the coefficients for standardization. After forecasting $z_f(t)$, we restore $\hat{z}_f(t)$ for the original time series. For $p=1$, we restore $\hat{y}_f(t+p|t)$ using $\hat{z}_f(t+p|t)$ as follows:
\begin{equation}
    \hat{y}_f(t+p|t) = \sigma_f \hat{z}_f(t+p|t) + \mu_f - \sum_{k=1}^{N-1} h_f(k) y_f(t+p-k).
\end{equation}
For $p \geq 2$, we obtain the $p$-steps-ahead forecast $\hat{y}_f(t+p|t)$ by substituting forecasts for the unobserved $y_f(t \geq t+1)$ as follows:
\begin{equation} \label{eq:restore_filter2}
    \hat{y}_f(t+p|t) = \sigma_f \hat{z}_f(t+p|t) + \mu_f \left[-\sum_{k=1}^{p-1} (h_f(k) \hat{y}_f(t+p-k|t)) + \sum_{k=p}^{N-1} (h_f (k) y_f (t+p-k)) \right].
\end{equation}
We can restore $\hat{y}_f(t+p|t)$ for any $p$ to apply equation~(\ref{eq:restore_filter2}) recursively.

We can apply FIR filters such as the moving average filter to suppress noise. In contrast, an existing study\cite{Okuno2017a} focuses on the filtered delay coordinates constructed by blending the derivatives of time series and the original one to search the nearest neighbors. Although these are a type of high-pass FIR filters, such filters improve the prediction accuracy in less noisy time series.

\subsection*{Overall algorithm of the proposed framework}
We denote a multivariate time series as $y(t) \in \mathbb{R}^n$ and the target variable to be forecasted as $y_f$. The proposed forecasting framework is presented in Algorithm~\ref{alg:framework} in terms of the user-defined number of split training datasets $K$, the number of top-performing embeddings chosen for each split $M$, the linear filters, and the number of filters $L$.

\bibliography{library}

\section*{Acknowledgments}
We thank Professor Christian W. Dawson for permission to use the flood dataset. This research is partially supported by Kozo Keikaku Engineering Inc., JSPS KAKENHI 15H05707, WPI, MEXT, Japan, AMED JP19dm0307009, and JST CREST JPMJCR14D2, Japan. We thank Dr. Takahiro Omi for fruitful discussions.

\section*{Author contributions statement}
S.O., K.A., and Y.H. conceived the original design of this study.
S.O. and A.K. designed the study of the toy models, and S.O. and Y.H. designed the study of the flood data.
S.O. conducted the numerical experiment.
All authors contributed to interpreting the results and writing the manuscript.
All authors checked the manuscript and agreed to submit the final version of the manuscript.

\section*{Additional information}
\textbf{Competing Interests}: The authors declare that they have no competing interests.

\begin{table}[hb]
    \centering
    \begin{tabular}{|l|l|l|l|l|l|l|}
        \hline
        Prediction steps [h] & Proposed       & RDE   & MVE            & LSTM  & SVR   & Random Forest \\
        \hline
        6      & 0.068          & 0.069 & \textbf{0.060} & 0.123 & 0.073 & 0.079  \\
        \hline
        12     & \textbf{0.167} & 0.174 & 0.171          & 0.213 & 0.177 & 0.185  \\
        \hline
        18     & \textbf{0.247} & 0.278 & 0.273          & 0.291 & 0.268 & 0.275  \\
        \hline
        24     & \textbf{0.349} & 0.389 & 0.373          & 0.372 & 0.365 & 0.363  \\
        \hline
    \end{tabular}
    \caption{\label{tab:flood}RMSEs of the flood dataset computed using the proposed framework, randomly distributed embedding (RDE), and multiview embedding (MVE) and also compared with those obtained by long short-term memory (LSTM), support vector regression (SVR), and the random forest. The proposed framework achieved the best accuracy for the 12-, 18-, and 24-h-ahead forecasts. Although the result obtained by the proposed framework for the 6-h-ahead forecast is the second best, the difference is slight.}
\end{table}



\begin{algorithm}[pth]
    \caption{Proposed forecasting framework}
    \label{alg:framework}
    \begin{algorithmic}[1]
        \State Initialize suboptimal embeddings $\mathcal{S}$ \Comment{First step: Obtaining suboptimal embeddings}
        \For {each $K'$ in $\{1,2,...,K\}$}
            \State Initialize a set of the Hall of Fame embeddings $\mathcal{H}$
            \State Initialize a set of embeddings $\mathcal{P}$ (population)
            \For {each generation}
                \State Compute $\hat{y}_f^e (t+p|t) \ \forall t \in \mathcal{T}_{train}^{K'}$ by map $\psi_e \ \forall e \in \mathcal{P}$
                \State Evaluate the fitness by $\sum_{t \in \mathcal{T}_{train}^{K'}} \sum_p \|\hat{y}_f^e (t+p|t) - y_f (t+p)\|$
                \State Append the elements of $\mathcal{P}$ to $\mathcal{H}$
                \State Create a new $\mathcal{P}$ for the next generation using an evolution strategy
            \EndFor
            \State Sort $\mathcal{H}$ by fitness
            \For {each $e_h$ in $\mathcal{H}$}
                \State \algorithmicif\ $d(e_h, e_s) \geq \theta \ \forall e_s \in \mathcal{S}$ \algorithmicthen\ Append $e_h$ to $\mathcal{S}$ with $L$ filters
                \State \algorithmicif\ $M \cdot L$ embeddings are appended to $\mathcal{S}$ \algorithmicthen\ break
            \EndFor
        \EndFor
        \For {each $e$ in $\mathcal{S}$} \Comment{Second step: Combining forecasts to minimize the in-sample error}
            \State Compute $\hat{y}_f^e (t+p|t) \ \forall t \in \mathcal{T}_{train}$ by map $\psi_e$
            \State Evaluate the in-sample error of $\hat{y}_f^e (t+p|t)$
        \EndFor
        \State $\mathcal{I}_p(i)$ := Sort $\mathcal{S}$ by the in-sample error of the $p$th step
        \For {each $k$ in the indices of $\mathcal{I}_p$}
            \State Compute the combined forecast $Y_k (t+p|t)$ by a recursive formula
            \State Evaluate the in-sample error $Err(k)$ of $Y_k (t+p|t)$
        \State Obtain the optimal number to combine $\hat{k}_p = \argmin_{k=1,2,...,P} \ Err$
        \EndFor
        \While {$t \geq 0$} \Comment{Forecast unobserved samples $t \geq 0$ by $\hat{k}_p$ embeddings}
            \For {$k=1$ to $\hat{k}_p$}
                \State Compute $\hat{y}_f^{\mathcal{I}_p(k)} (t+p|t)$
            \EndFor
            \State Yield the combined forecast by $\sum_{k=1}^{\hat{k}_p} \hat{y}_f^{\mathcal{I}_p(k)}(t+p|t) / \hat{k}_p$
        \EndWhile
    \end{algorithmic}
\end{algorithm}

\end{document}


\flushbottom
\maketitle
\thispagestyle{empty}
\section*{Effect of multiple optimizations}
We solve multiple combinatorial optimization problems to divide the objective function into $K$ problems in the first step.
This procedure yields diverse embeddings to improve the performance of the combined forecast.
Here, we numerically show the effect of the multiple optimizations.

We profiled the effect of $K$ times optimization with an example of the 10-dimensional Lorenz'96I model with the data length of 4000.
We forecasted the data by two methods: one is the proposed $K$ optimizations, and the other is a single optimization to minimize the whole training in-sample error.
All other processes follow the proposed forecast framework.
Note that the computational time is almost equivalent with both methods.

\begin{figure}[ht]
    \centering
    \includegraphics[width=1.\linewidth]{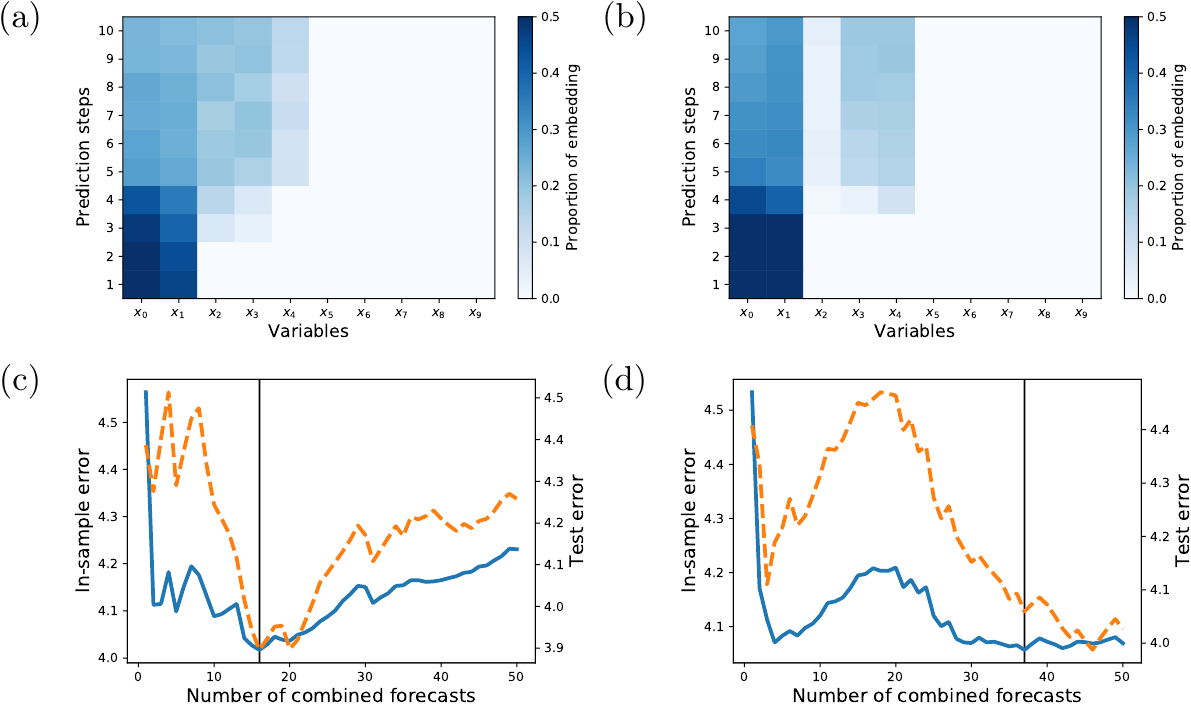}
        \caption{Effect of multiple optimizations: the proportion of embedding of the proposed forecasts with (a) $K$ times optimization and (b) a single optimization to minimize the whole in-sample error. The color indicates the proportion of embedding for each variable averaged over the number of combined forecasts for each step. Panels (c) and (d) show the relation between the number of combined forecasts and the five-steps-ahead errors for (c) the $K$ times optimization and (d) the single optimization, respectively. The solid line shows the in-sample error, and the dashed line shows the test error. The vertical solid line shows the selected number of combined forecasts.}
        \label{fig:kdiv}
\end{figure}

As shown in Figures~\ref{fig:kdiv}(a) and (b), the procedure with $K$ optimizations enhanced the diversity of embeddings compared with the single optimization.
This result suggests that the $K$ times optimization found different but useful embeddings for each optimization process.
The $K$ times optimization also reduced the number of forecasts to combine (Figures~\ref{fig:kdiv}(c) and (d)).
This is because the single optimization process yielded similar embeddings, and the combination of similar forecasts yields a small performance improvement empirically\cite{Sollich96, Kuncheva2003}.
Reductions in the number of combined forecasts and the computational time are important, especially for real-time applications.

\section*{Application to low-dimensional dynamics}
We applied the proposed forecast framework to low-dimensional datasets---namely, the Lorenz'63 equations\cite{Lorenz1963} (three variables), the R\"ossler equations\cite{Rossler1976} (three variables), and the six-dimensional Lorenz'96I equations\cite{Lorenz1995a} (six variables).
We set the data length to 4000 as the database and evaluated forecasts up to 10 steps ahead with 500 samples for all the cases. See the next section for the detailed conditions.

\begin{figure}[ht]
    \centering
    \includegraphics[width=1.\linewidth]{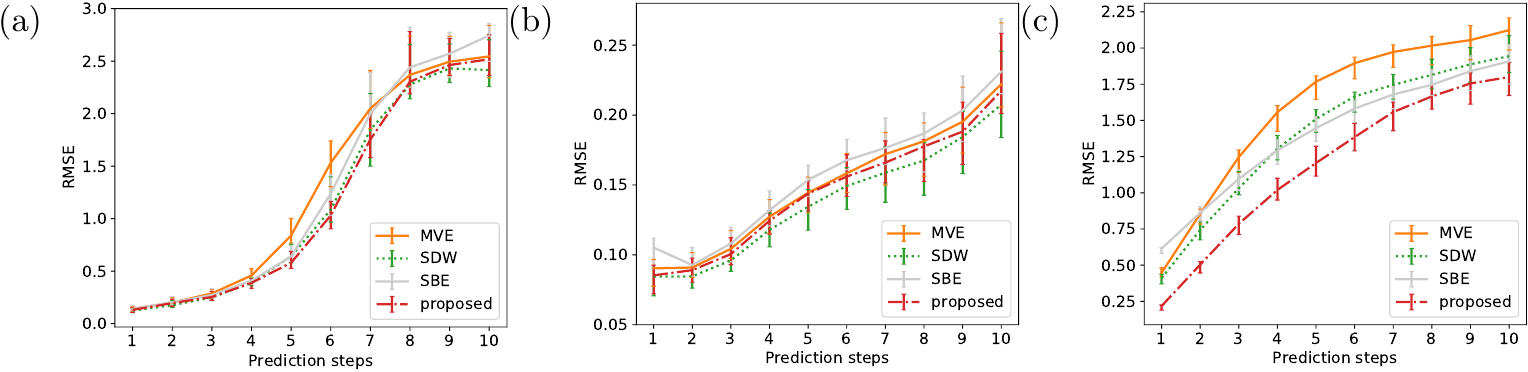}
    \caption{Forecast performance for low-dimensional datasets: RSMEs for (a) the Lorenz'63 dataset, (b) the R\"ossler dataset, and (c) the six-dimensional Lorenz'96I dataset. We compared the performance of multiview embedding (MVE), state-dependent weighting (SDW), and single-best embedding based on the $(\mu+\lambda)$-ES algorithm (SBE). These tests were carried out with 20 datasets generated with different random initial conditions and noise. The median, upper quartile, and lower quartile are shown.}
    \label{fig:lowdims}
\end{figure}

As shown in Figure~\ref{fig:lowdims}, the proposed framework did not always achieve the best performance. There is little difference between the proposed framework and state-dependent weighting (SDW) for the Lorenz'63 dataset (Figure~\ref{fig:lowdims}(a)), and SDW achieved the best performance for the R\"ossler dataset (Figure~\ref{fig:lowdims}(b)).
In contrast, when we increased the number of variables to six, the proposed framework yielded the best performance (Figure~\ref{fig:lowdims}(c)), and as stated in the main text, the proposed framework achieved much better results than the others for cases with 10 or 20 variables.

Interestingly, the performance of SDW was superior to that of MVE for all of the low-dimensional toy models mentioned above, although the performance of SDW for the high-dimensional dataset was not satisfactory, as stated in the main text. This result means that weighting based on the forecast performance and its corresponding state\cite{Okuno2019} worked fine with simple low-dimensional data.
On the other hand, it is difficult to improve the performance by weighting for high-dimensional data, and a simple average is sufficient for such cases.
In short, it is preferable to apply the proposed framework to complex high-dimensional and less noisy data, but it is worth considering the application of weighting methods such as SDW to simple low-dimensional data.

\section*{Detailed conditions of the numerical experiments}
\subsection*{Lorenz'96I equations}
The Lorenz'96I equations\cite{Lorenz1995a} are expressed as 10-dimensional differential equations as follows:
\begin{equation}
    \frac{d x_i}{d t} = x_{i-1} (x_{i+1}-x_{i-2}) - x_i + F,
    i=0,1,2,...,9,
\end{equation}
where $F$ is a forcing variable and $i$ is cyclic. We generated 10-dimensional time series: Lorenz'96I's $x_0, x_1, x_2, x_3$, and $x_4$ and random walks $x_5, x_6, x_7, x_8$, and $x_9$. We chose the initial condition from a normally distributed random value for each variable. We set the integration step to 0.001 and recorded every 50 points. Note that the initial transients were disregarded. We forecasted $x_0$ up to 10 steps. We set $K=10, M=3, \theta=3$, and $h_i(0)=1, h_i(1)=\rho \ \forall i$ for $\rho \in \{0.0, -0.2, -0.4, -0.6, -0.8, -1.0\}$. For the $(\mu+\lambda)$-ES algorithm, we set $\mu=50, \ \lambda=100$, the number of generations to 10, and the number of populations to 100.

We compared the proposed framework with existing frameworks---namely, randomly distributed embedding with an aggregation scheme (RDE), multiview embedding (MVE), SDW with $E=4$, and up to four lags for MVE and SDW. Because the possible number of embeddings combinatorially increases, we randomly generated 1000 embeddings instead of using brute-force calculation.
Note that the total number of embedding evaluations is almost the same as that for the proposed framework.
We also computed the single-best embedding via the $(\mu+\lambda)$-ES algorithm to minimize the total in-sample error within the whole training dataset.

\subsection*{Kuramoto--Sivashinsky equations}
The Kuramoto--Sivashinsky equations\cite{Kuramoto1976, Sivashinsky1977} are expressed as follows:
\begin{equation}
    \frac{\partial y}{\partial t} = -\frac{\partial^2 y}{\partial x^2} - \frac{\partial^4 y}{\partial x^4} - u \frac{\partial y}{\partial x}.
\end{equation}
We computed $y(x,t)$ with spatially periodic boundary conditions in the interval $[0, L]$, where $L=22$ and the number of uniform grids is 128. We generated 20-dimensional time series: the values of the first 10 grids of the Kuramoto--Sivashinsky equations $x_0, x_1, ..., x_9$ with a sampling time of 1.0 and 10 random walks $x_{10}, x_{11}, ..., x_{19}$.
We set $E=5$ and considered up to five lags for the existing schemes (RDE, MVE, and SDW).
We forecasted $x_0$ up to 10 steps using the same parameter values, data length, and noise scales as those in the Lorenz'96I example.

\subsection*{Flood dataset}
The flood forecasting competition dataset ``Artificial Neural Network Experiment (ANNEX 2005/2006)''\cite{Dawson2005a} contains river stage and rainfall data for three periods: 1993-10-01 to 1994-03-31, 1994-10-01 to 1995-03-31, and 1995-10-01 to 1996-03-31. We forecasted river stage $Q$ 6, 12, 18, and 24 h ahead using nine variables: the river stages of the target ($Q$) and three upstream sites ($US1$, $US2$, and $US3$) and five rain gauges ($RG1$, $RG2$, $RG3$, $RG4$, and $RG5$). All data were sampled by 6 h. We used the period of 1994-10-01 to 1995-03-31 for testing and the others for training. We set $K=6, M=3, \theta=3$, and $h_i(0)=1, h_i(1)=\rho \ \forall i$ for $\rho \in \{0.0, -1.0\}$. For the $(\mu+\lambda)$-ES algorithm, we set $\mu=50, \lambda=100$, the number of generations to 20, and the number of populations to 100.

\subsection*{Lorenz'63 equations}
The Lorenz'63 equations\cite{Lorenz1963} are expressed as three-dimensional differential equations as follows:
\begin{equation}
    \frac{dx}{dt} = p (y - x),
    \frac{dy}{dt} = x (r - z) - y,
    \frac{dz}{dt} = xy - bz.
\end{equation}
We generated three-dimensional time series: Lorenz'63's $x$ and $y$ and a random-walk series. We set the integration step to 0.001 and recorded every 100 points with $p=10,b=8/3,$ and $r=28$. Note that the initial transients were disregarded. We forecasted $x$ up to 10 steps. We set $K=10, M=3, \theta=3$, and $h_i(0)=1, h_i(1)=\rho \ \forall i$ for $\rho \in \{0.0, -0.2, -0.4, -0.6, -0.8, -1.0\}$. For the $(\mu+\lambda)$-ES algorithm, we set $\mu=50, \lambda=100$, the number of generations to 10, and the number of populations to 100.

We compared the proposed framework with MVE and SDW with $E=4$ up to five lags and the single-best embedding via the $(\mu+\lambda)$-ES algorithm to minimize the total in-sample error within the whole training dataset. Note that we did not carry out calculations with RDE because we were not able to prepare a sufficient number of ``nondelay embeddings'' for this type of low-dimensional data.

\subsection*{R\"ossler equations}
The R\"ossler equations\cite{Rossler1976} are expressed as three-dimensional differential equations as follows:
\begin{equation}
    \frac{dx}{dt} = -y - z,
    \frac{dy}{dt} = x + ay,
    \frac{dz}{dt} = b + z (x-c).
\end{equation}
We generated three-dimensional time series: R\"ossler $x$ and $y$ and a random-walk series. We set the integration step to 0.001 and recorded every 500 points with $a=0.36,b=0.4$, and $c=4.5$. Note that the initial transients were disregarded. We forecasted $x$ up to 10 steps with the same conditions as those of the Lorenz'63 equations.

\subsection*{Six-dimensional Lorenz'96I equations}
The six-dimensional Lorenz'96I dataset contains the six-dimensional Lorenz'96I series $x_0, x_1$, and $x_2$ and random walks $x_3, x_4$, and $x_5$. The conditions for numerical integration were the same as those of the 10-dimensional Lorenz'96I equations. We forecasted $x_0$ up to 10 steps with the same conditions as those of the Lorenz'63 equations.

\section*{Method of analogues}
We applied a variation of the method of analogues\cite{Lorenz1969} to obtain a $p$-steps-ahead forecast $\hat{y}_f (t+p|t)$ at time $t$. The method of analogues takes the forward paths of neighboring trajectories as the forecast. Here, we forecast $y_f (t+p|t)$ from the set of delay coordinates $\{v(t) \mid t \in \mathcal{T}_{train}\}$.
We first search the database to find the neighboring points of $v(t)$. Then, the method of analogues gives $\hat{v}(t+p|t)$ in terms of the set of neighboring time indices $\mathcal{I}(t)$ as follows:
\begin{equation}
    \hat{v}(t+p|t) = \sum_{t' \in \mathcal{I} (t)} \lambda(t') v(t'+p),
\end{equation}
where $\lambda(t') \in \mathbb{R}_{+}$ is a weight satisfying $\sum_{t' \in \mathcal{I}(t)} \lambda (t)$. We employ $\lambda (t') \propto \|v (t') - v (t)\|^2_2$ throughout this paper.

Note that we can apply more advanced methods such as that in Refs.~\citenum{Farmer1987, Hirata2014} or any other regression methods with our proposed forecasting framework.


\bibliography{library}